\documentclass[10pt,twocolumn,letterpaper]{article}
\usepackage{btas}
\usepackage{times}
\usepackage{epsfig}
\usepackage{graphicx}
\usepackage{amsmath}
\usepackage{amssymb}
\usepackage[ruled,vlined]{algorithm2e}
\usepackage{color}
\usepackage[numbers,sort&compress]{natbib}
\usepackage[norule,symbol,perpage]{footmisc}

\newcommand{\imgStub}[3]{\begin{minipage}{#1\textwidth}\begin{center}
       \includegraphics[width=1\textwidth]{#2}\\
#3\end{center}\end{minipage} \hfil}

\btasfinalcopy % *** Uncomment this line for the final submission

\ifbtasfinal\fi

\makeatletter 
\def\ps@IEEEtitlepagestyle{ 
\def\@oddfoot{\mycopyrightnotice} 
\def\@evenfoot{} 
} 
\def\mycopyrightnotice{ 
{\hfill \footnotesize 978-1-7281-1522-1/19/\$31.00 \copyright 2019 IEEE\hfill} 
} 
\makeatother 

\begin{document}

%%%%%%%%% TITLE
\title{A Genetic Algorithm Enabled Similarity-Based Attack on Cancellable Biometrics}

\author{Xingbo Dong$^{\dagger}$,  Zhe Jin$^{\dagger}$, Andrew Teoh Beng Jin$^{\ddagger}$\\
$^{\dagger}$School of Information Technology, Monash University Malaysia, Subang Jaya, 46150, Malaysia \\
$^{\ddagger}$School of Electrical and Electronic Engineering,Yonsei University, Seoul, South Korea\\
{\tt\small \{xingbo.dong,jin.zhe\}@monash.edu, bjteoh@yonsei.ac.kr}
}
\maketitle
\thispagestyle{empty}

\begin{abstract}
Cancellable biometrics (CB) as a means for biometric template protection approach refers to an irreversible yet similarity preserving transformation on the original template. With similarity preserving property, the matching between template and query instance can be performed in the transform domain without jeopardizing accuracy performance.  Unfortunately, this trait invites a class of attack, namely similarity-based attack (SA). SA produces a preimage, an inverse of transformed template, which can be exploited for impersonation and cross-matching. In this paper, we propose a Genetic Algorithm enabled similarity-based attack framework (GASAF) to demonstrate that CB schemes whose possess similarity preserving property are highly vulnerable to similarity-based attack. Besides that, a set of new metrics is designed to measure the effectiveness of the similarity-based attack. We conduct the experiment on two representative CB schemes, i.e. BioHashing and Bloom-filter. The experimental results attest the vulnerability under this type of attack.
\end{abstract}
\let\thefootnote\relax\footnotetext{\mycopyrightnotice} 
\section{Introduction}

Human biometrics as identity credential offers great usability for identity management. Biometrics based authentication systems have been widely deployed and commonplace. With the prevalent of biometrics systems, the public are getting concern about the security and privacy of the biometric templates if they are compromised. For instance, if an adversary manages to retrieve an individual's template, impersonation with the stolen template is straightforward. Furthermore, biometrics is intrinsically linked to individual and it is limited, thus revocation and replacement like passwords are impossible. With the advancement of technology, another critical threat on privacy is that the original biometric data can actually be reconstructed with high accuracy ~\cite{mai2018reconstructiondeepface,galbally2013irisreconstruct,cao2015learningfingerreconstruct}. 

Due to above concern, cancellable biometrics (CB) has been devised for biometric templates. CB is a parameterized irreversible yet revocable transform to ensure security and privacy of the biometric template.~\cite{patel_cancelable_2015-BTPOVERVIEW,sandhya_biometric_2017-BTPOVERVIEW,chandra_cancelable_2011-BTPOVERVIEW,nandakumar_biometric_2015-BTPOVERVIEW}. Revocability of CB demands the transformed template can be cancelled and replaced by changing the parameter (or known as helper data) whenever required. The transformed templates should be independent to each other as well as with original template to prevent cross-matching (unlinkability criterion). This can be done with application-specific helper data. Another vital criterion is irreversibility wherein the original template should be computationally difficult be inverted from a single instance or multiple instances of transformed templates with or without helper data.

Numerous CB schemes have been proposed in the literature and shown satisfy the CB criteria up to certain level. Several specific attacks have also been proposed for CB schemes such as brute-force inversion attack, dictionary attack (or known as zero effort false accept attack),  correlation attack (or known as attacks via record multiplicity, ARM) and similarity-based attack (also named masquerade or preimage attack).  (see review papers \cite{patel_cancelable_2015-BTPOVERVIEW,sandhya_biometric_2017-BTPOVERVIEW,chandra_cancelable_2011-BTPOVERVIEW,nandakumar_biometric_2015-BTPOVERVIEW}). While correlation attack is associated to unlinkability criterion, dictionary attack exploits decision threshold of biometric systems and brute-force inversion is related to irreversibility, similarity-based attack breaches all the criteria mentioned above.   

Brute-force inversion assumes by given transformed templates with or without helper data, an adversary may obtain an exact or approximate original template. The attack could be largely resisted if the transformation is a many-to-one function~\cite{ratha2007transformation} or salted~\cite{jin_TIFS_2018-IoM}. On the other hand, similarity-based attack (SA) attempts to find a preimage, an inverse of the transformed template, which may or may not close to original template that against irreversibility criterion. Furthermore, an attacker can leverage the preimage with the compromised helper data to impersonate the genuine user even his/her template is protected. Accordingly, cross-matching also become possible when the associated application-specific helper data is known. SA is enabled by the similarity preserving property of the cancellable biometric transformation while this property is essential to achieve non-degradation in terms of accuracy performance after transformation. To be specific, the distance or similarity between templates in the original space should preserve relatively in its transform domain. Hence, SA utilizes the distances relationships between templates instead of the exact distances between them for inversion~\cite{lacharme2013preimagebiohashingga,chen2019deepDSQ}. In literature, SA is often ignored while the rest of the attacks have became a must component to overcome wherever a new CB scheme is designed~\cite{pillai-2011-secure,jin_TIFS_2018-IoM} .

There are a few realizations of SA. In 2010, Nagar et al. propose a method to find preimage based on a given BioHashing template (A salting-based CB scheme) and its parameters~\cite{nagar2010biometricsecurityanalysis}. The Biohashing is formulated as an (approximated) linear equation system and the reconstruction problem is solved via a bounded least-squares solver. However, this method does not work on non-linear systems and sensitive to outlier ~\cite{natrella2010nist}.

In 2012, Feng et al. propose a two-step method to retrieve preimage from binary transformed templates~\cite{feng2014masquerade} . In the first step, given a binary template $b$, the bases $w_p$, and the threshold $t_p$, the reconstruction to real-valued templates problem can be solved by using a Perceptron. In the second step, a modified hill-climbing algorithm is used to construct the fake face image from the real-valued template.  Despite this scheme can achieve good performance, the implementation is complex. The latest work about SA is published in 2019~\cite{chen2019deepDSQ} . The authors justify theoretically most of the existing CB schemes are highly susceptible to SA as they largely rely on the similarity preserving property to maintain accuracy performance. However, no implementation is given in this work.  

Despite a handful works of SA has been proposed, they are either method or biometric modality dependence or of inefficient, such as hill climbing that requires long iterations. In this paper, we introduce a Genetic Algorithm based similarity-based attack framework (GASAF) that works efficiently, effectively and easily irrespective to CB algorithm or biometric modality. With this framework, we break two well-known cancellable schemes, namely BioHashing~\cite{teoh_random_2006} on deep model face features and Bloom-filter~\cite{rathgeb2013bloomfilteriris} on IrisCode~\cite{daugman2006iriscode}. We made our source code available at https://bit.ly/2FduSYl. 

The contributions of this paper are as follows:
\begin{itemize}
\item We propose a framework of similarity-based attack by means of GA (GASAF). The GASAF can be applied to CB schemes that transformed real-valued and binary biometrics templates. The former instance is face feature vector that generated by the deep network and the latter is IrisCode.
\item We demonstrate the GASAF can find the preimage from its transformed counterparts in an efficient and effective manner. 
\item We design a set of new metrics to evaluate the effectiveness of the similarity-based attack for cancellable biometrics.
\end{itemize}

%%%%%%%%% BODY TEXT
\section{Background}
In this section, we first provide a brief account of Locality Sensitive Hashing (LSH) in which most of the CB schemes based upon and then followed by the description of BioHashing and Bloom-filter.

\begin{figure}[t]
\begin{center}
\includegraphics[width=0.95\linewidth]{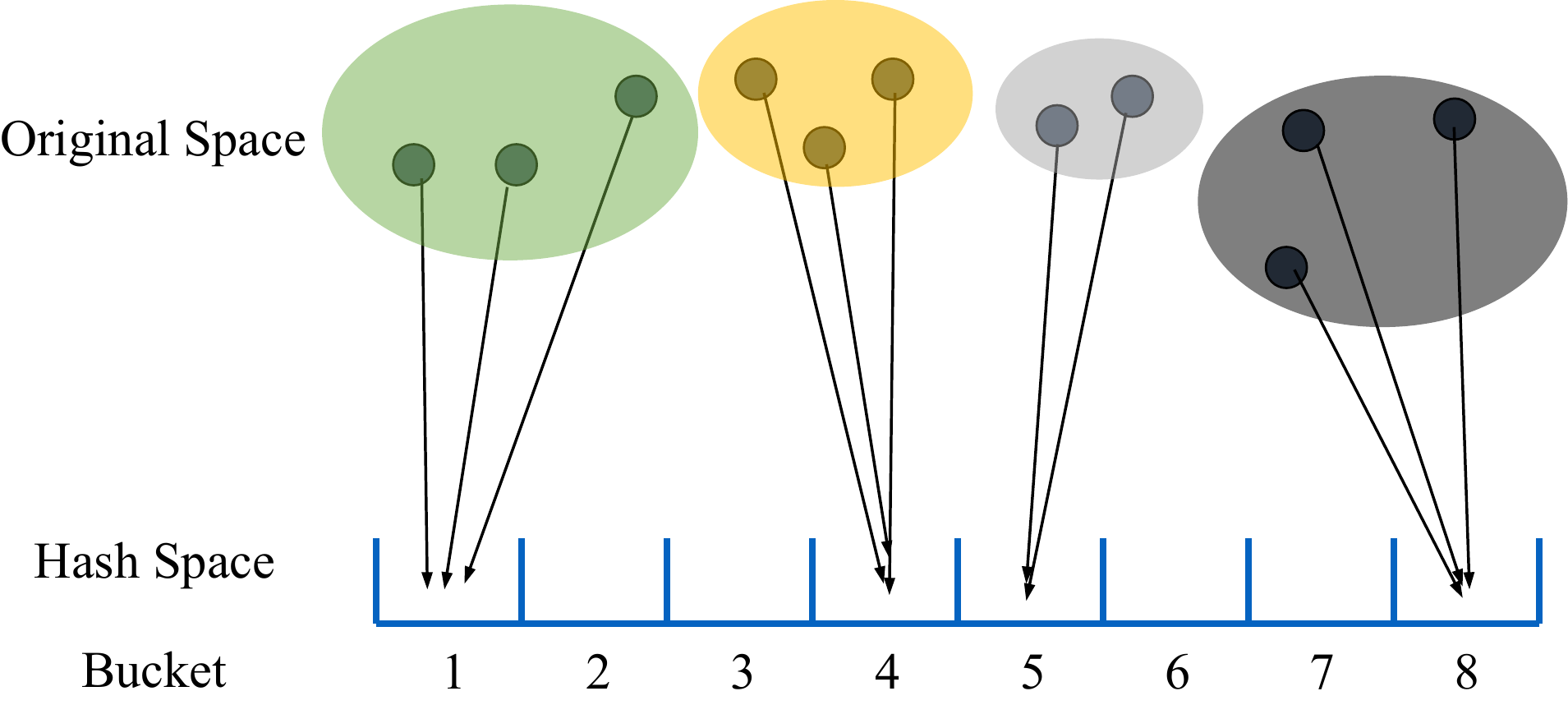}
\end{center}
  \caption{Locality sensitive hashing (LSH)\label{fig:lsh}}
\end{figure}

Cancellable biometrics schemes leverage similarity preserving property to enable accuracy performance preservation. More specifically, given two biometric feature sets $X,Y \in \mathbb{R}^d$ in its original feature space, they are transformed into a new space as $X^\prime,Y^\prime \in \mathbb{R}^m$ , and the distance between $X $ and $ Y$ should be nearly preserved in the new space $\mathbb{R}^m$. Therefore, matching can be accomplished in the transformed domain. LSH is one such a notion used as a transformation function in CB attributed to its similarity/distance preserving characteristic. LSH aims to map similar items into same "buckets" with a maximized probability (see Figure \ref{fig:lsh} ).

Given the LSH family $\mathcal{H}=\{h_i: \mathbb{R}^d \to S\}$ which maps data points from $\mathbb{R}^d$ to a bucket $s\in S$ , and the similarity function $d(\cdot)$, for any two given points $X,Y \in \mathbb{R}^d$,the LSH family satisfies the following conditions :
\begin{equation}
\begin{split}
\mathbb{P}_{h\in \mathcal{H}} (h (X)=h (Y))\leq \gamma,if~d(X,Y) < \alpha \\
\mathbb{P}_{h\in \mathcal{H}} (h (X)=h (Y)) \geq\delta,if~d(X,Y)> \beta
\end{split}
\end{equation}
where \begin{math}\delta>\gamma\end{math}. 
 
BioHashing is a representative salting-based generic CB scheme~\cite{teoh_random_2006}. It is a two-factor scheme based on the user-specific token, act as salt (helper data) and real-valued biometric feature vector, and followed by a binarization procedure. The $n$-bit BioHash code $c$ of a feature vector $x\in R^N$ is computed as $c=Sgn(\sum xb_i-\tau)$, where $Sgn(\cdot)$ is a signum function, and $\tau$ is an empirically determined threshold, and $b_i\in R^N,i=1,...,n (n\leq N) $ is a random vector. The Hamming distance is computed between two hash codes to indicate the similarity between two biometric vectors. New instance of $x$ can be reissued by replacing a newly generated pseudo-random vector.

The theory of Biohashing is grounded in random projection (RP), which is an instance of LSH. RP is a process of projecting feature vector from $n$ dimensions to $m$ dimensions ($n \gg m$) in the Euclidean space by using random matrices~\cite{pillai-2011-secure}. RP is based on Johnson-Lindenstrauss lemma (J-L lemma)~\cite{beals_1984_extensionsJLlemma} that warrants the points from a high-dimensional space can be embedded onto low-dimensional space while preserving the distance approximately.

Bloom-filter is adapted as a means of generic CB schemes and also shares same characteristic of LSH~\cite{gomez-2014-protected}. In bloom-filter, the biometric feature is mapped to a bit array $b$ with several independent hashing functions, where $b$ is an bit-array of length $n$, $b\in {[0,1]}^n$. Specifically, $k (k \ll n)$ independent hash functions denoted as $h_1,h_2,...,h_k$ are pre-defined first, then each element in set $S$ is hashed and the hashed result is derived as $k$ indices. Finally, set all $k$ indices of the bit-array $b$ to unity. At the verification stage, the bit-array of the query element $y$ is matched to the stored template by means of hamming distance. The Bloom-filter is proposed on IrisCode by Rathgeb et. al~\cite{rathgeb2013bloomfilteriris}. In their approach, the $W\times H$ IrisCode is divided into $K$ blocks where each block consists of $l=W/K$ columns of codewords. The word size of each codeword is denoted as $\omega$ bits. Finally $K$ Bloom-filters with length of $2^\omega$ bits are generated as the transformed template. 

\section{Similarity-Based Attack via Genetic Algorithm}

In this section, a general framework of SA is presented firstly, followed by a detailed implementation based on Genetic Algorithm. Finally the evaluation metrics of SA is proposed. 

\subsection{Framework of Similarity-Based Attack}
In SA, we introduce Kerckhoffs's assumption and assume that the attacker can access to the protected template, and known well about the transformation function as well as the parameters of the function.
The essential part of the similarity-based attack is to generate the original template approximately (preimage) based on an initial guess of template instance. SA can be launched by solving the following minimization problem:
\begin{equation}
\operatorname*{arg\,min}_{\hat{x}} \lVert  x-\hat{x}\lVert,	\label{eq.finalgoal}
\end{equation}
where $x$ is the original template and $\hat{x}$ is the pre-image of $x$. Given a CB transformation function $h(\cdot)$, the relative distance of two transformed templates should be preserved with respect to their counterparts in the original space, hence (\ref{eq.finalgoal}) can be rewritten as:
\begin{equation}
\operatorname*{arg\,min}_{\hat{x}} d(h(\hat{x}),h(x)),	\label{eq.finalgoal}
\end{equation}
where $d(\cdot)$ indicates an algorithm-specific distance function and $h(x)$ is the compromised template in the database. The cost function of the optimization problem can be defined as:
\begin{equation}
s\approx d(h(\hat{x}),h(x)), \label{eq.fitness}
\end{equation}
It is worth noting that the $h(\cdot)$ is the CB transformation function that possess similarity preserving property. The $x$ can be manifested in the form including discrete (binary or integer) and continuous-valued. 

As suggested by~\cite{kaplan2017known} and~\cite{chen2019deepDSQ}, $\hat{x}$ can be estimated by a search algorithm based on the similarity preserving characteristic of their transformed counterparts. Basically the process of estimating the best solution resembles number guess game wherein several steps are taken to 'guess' the best $\hat{x}$: 
\begin{enumerate}
\item Define the solution space and generate the first guess randomly; 
\item Project the guessed $\hat{x}$ into the transform space with $h(\cdot)$; 
\item Compute the cost of the guessed $\hat{x}$ by its cost function~(\ref{eq.fitness}); 
\item Generate a new guess based on the current guess and compute the cost. Repeat step 2 to step 4 until the stop condition is met. 
\end{enumerate}

\subsection{Genetic Algorithm \label{section.ga}}

GA is a commonly used robust search algorithm that simulates the genetic mechanism of the biology world such as mutation, crossover and selection. Like the biological evolution, GA repeatedly modifies a population of individual solutions to drive the population to evolve into a optimal solution. The children generation generally can be formed by three types of manipulations:
\begin{itemize}
\item Selection: select the individuals of generated features as parents, which is used to generate the next generation of features. 
\item Crossover: combine or exchange some part of two parents features to form children for the next generation of features, e.g. exchange some bits between two generated IrisCode.
\item Mutation: apply random changes to individual parents features to form children, e.g. inverse some bits of one generated IrisCode.
\end{itemize}
Specifically, the solution space encoding is generated firstly, and the first population is initialized randomly.A fitness function is designed to evaluate how optimal the individual in the population is, and the aforementioned three types of manipulation are used to form the next generation. Then same fitness function is used to evaluate the individual in the children population and continue the above procedures repeatedly (see Algorithm \ref{algo.gasaf}). Over successive generations, GA drives the population to evolve into a optimal solution. Until the newly generated individuals $\hat{x}$ reach the threshold of the fitness function or the average change in the fitness value less than certain threshold, the best individual is returned by GA. GA can be used to solve various optimization problems even though the objective function is discontinuous, non-differentiable, stochastic, or highly nonlinear. 

\begin{algorithm}[t]
\SetAlgoLined
\begin{flushleft}
    \textbf{INPUT:} CB transform function $h(\cdot)$, compromised template $h(x)$, fitness function $s(\cdot)$\\
    \textbf{OUTPUT:} Best individual of $\hat{x}$
\end{flushleft}
 Solution space encoding and population initialization\;
 Evaluate initial individuals by fitness function\;
 \While{!stopCondition}{
  Selection of the best-fit individuals for next generation\;
  Generate new child by crossover and mutation operations\;
  Evaluate new individuals by fitness function according to (\ref{eq.fitness})\;
  Replace the least-fit population with new individuals\;
 }
 \caption{Genetic Algorithm Based Similarity-Based Attack \label{algo.gasaf}}
\end{algorithm}

\subsection{Similarity-Based Attack Evaluation Metrics}
The key idea of SA is to find the preimage $\hat{x}$ of the original template $x$ based on the compromised template $h(x)$. Let $T_1$ be the helper data that associated to $h(x)$ in an application. We denote this process as $\hat{x} \gets \{h(x),T_1\} $. With $\hat{x}$, the attacker can gain access to other applications if $\hat{x}$ satisfies $d(h(\hat{x},T_2),h(x,T_2))<\theta$, where $d(\cdot)$ is a distance function and $\theta$ is the system threshold. 

\begin{figure}[t]
\begin{center}
\includegraphics[width=0.95\linewidth]{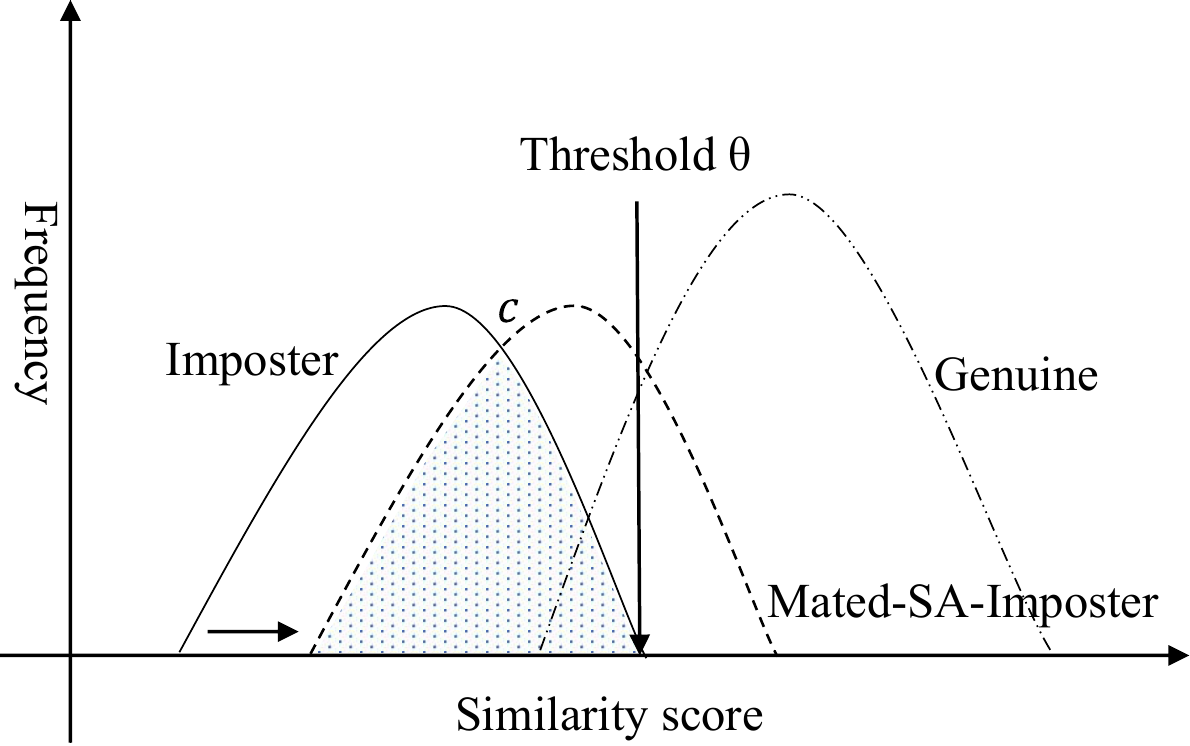}
\end{center}
  \caption{Score distribution shifting\label{fig:score_shift}}
\end{figure}

 For clarity, the matching scores under a few scenarios are defined as follows:

\begin{itemize}
\item \textbf{Genuine scores}: In a biometric authentication system, the similarity matching score between two transformed templates from same identity by same helper data is named as Genuine score;
\item \textbf{Imposter scores}: Matching score from different identity by same helper data is known as Imposter scores;
\item \textbf{Mated-SA-Imposter scores}: In the proposed SA, an attacker attempts to estimate the user original template, then use the preimage $\hat{x}$ to access another application. Under this situation, the matching score computed among $h(\hat{x})$ and its corresponding transformed template in the target database, i.e. $s=1-d(h(\hat{x}),h(x))$, is named as Mated-SA-Imposter scores (see Figure \ref{fig:score_shift}). 
\end{itemize}

Normally, the threshold value $\theta$ in Figure \ref{fig:score_shift} should be chosen to minimize the False Accept Rate (FAR) of the system. However, under SA, the Mated-SA-Imposter score distribution can 'shift' to the right as illustrated in Figure \ref{fig:score_shift}. This shift implies more matching scores will exceed threshold thus become easier to gain access to the system. 

Based on the above discussion, the overlapping region of the Imposter and Mated-SA-Imposter score distributions, denoted as $OL$, is used to evaluate the system's resistances to SA. If $OL=1$, this indicates the Mated-SA-Imposter and Imposter are overlapped totally, hence the attacker would fail to gain access. Otherwise, $OL=0$ implies the attacker can easily gain access with $\hat{x}$ since the matching score is greater than the threshold. 

To compute the overlapping region of two distributions, we assume the score distribution follows Gaussian distribution for simplistic. Let $S_1$ be the Imposter score and $S_2$ be the Mated-SA-Imposter score, hence $S_1\sim N( \mu_1,\delta_1 )$, $S_2\sim N( \mu_2,\delta_2 )$ where $\mu_1<\mu_2$. Let $c$ denote the intersection point of the overlapping region, the area of the intersection zone can be computed as:
\begin{equation}
OL=P(S_1>c)+P(S_2<c).
\end{equation}
Even though $OL$ can be used to indicate the score distribution shifting, the overlapping area depends on the specific score distribution, hence may not reflect the exact shifting circumstance. Hence apart from the overlapping of the distribution (shifting), a more practical indicator - false accept rate (FAR) under the pre-defined threshold is also used to measure the system's resistance to SA. The pre-defined threshold in this work is determined based on the matching score corresponding to the Equal Error Rate (EER) of the Genuine and Imposter scores. The Mated-SA-Imposter scores is used to compute the FAR of SA under threshold of EER, and here we denote as FAR@ET (FAR at EER threshold).

\section{Experiments}
To validate GASAF, two representative CB schemes are employed. BioHashing is tested firstly on a face benchmark dataset, i.e. LFW. Then Bloom-filter is evaluated on an iris dataset, i.e. CASIA-Iris-Interval. 

To explore the capability of successive attack for BioHashing, LFW~\cite{huang-2008-LFW} is used in this work to compute the above scores distribution under different simulations.Since GA is time consuming, and LFW contains 5749 subjects, it is impossible to perform the SA test on all LFW subjects, hence we build a customized LFW dataset to validate our work. In LFW, the subjects who have more than 10 images are selected and the first 10 images are choose to form a new small dataset denoted as LFW10 (158 users). The face feature of LFW10 is extracted by a deep network namely InsightFace (a.k.a ArcFace)~\cite{deng2018arcface}. 

As for Bloom-filter, the left eye images in the CASIA-v4-Iris-Interval (The Center of Biometrics and Security Research, http://www.cbsr.ia.ac.cn/china/Iris Databases CH.asp) are used to generate the IrisCode~\cite{daugman2006iriscode} by Libor Masek's method~\cite{LiborMasekiriscode}. 

\subsection{BioHashing}
\begin{figure*}[t!]
\setlength{\abovecaptionskip}{8pt} 
\setlength{\belowcaptionskip}{-14pt} 
\begin{center}
  \imgStub{0.32}{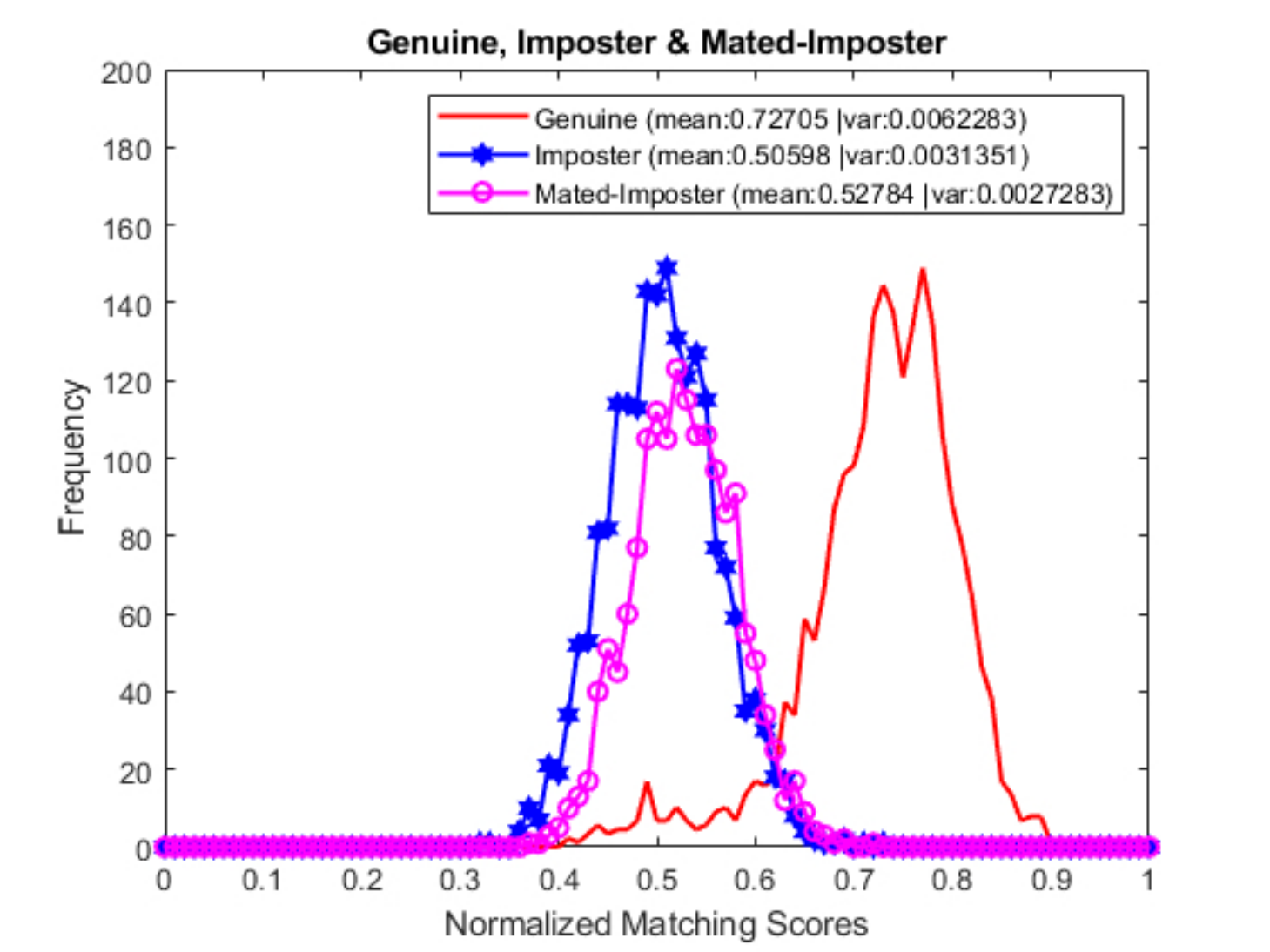}{(a) }
  \imgStub{0.32}{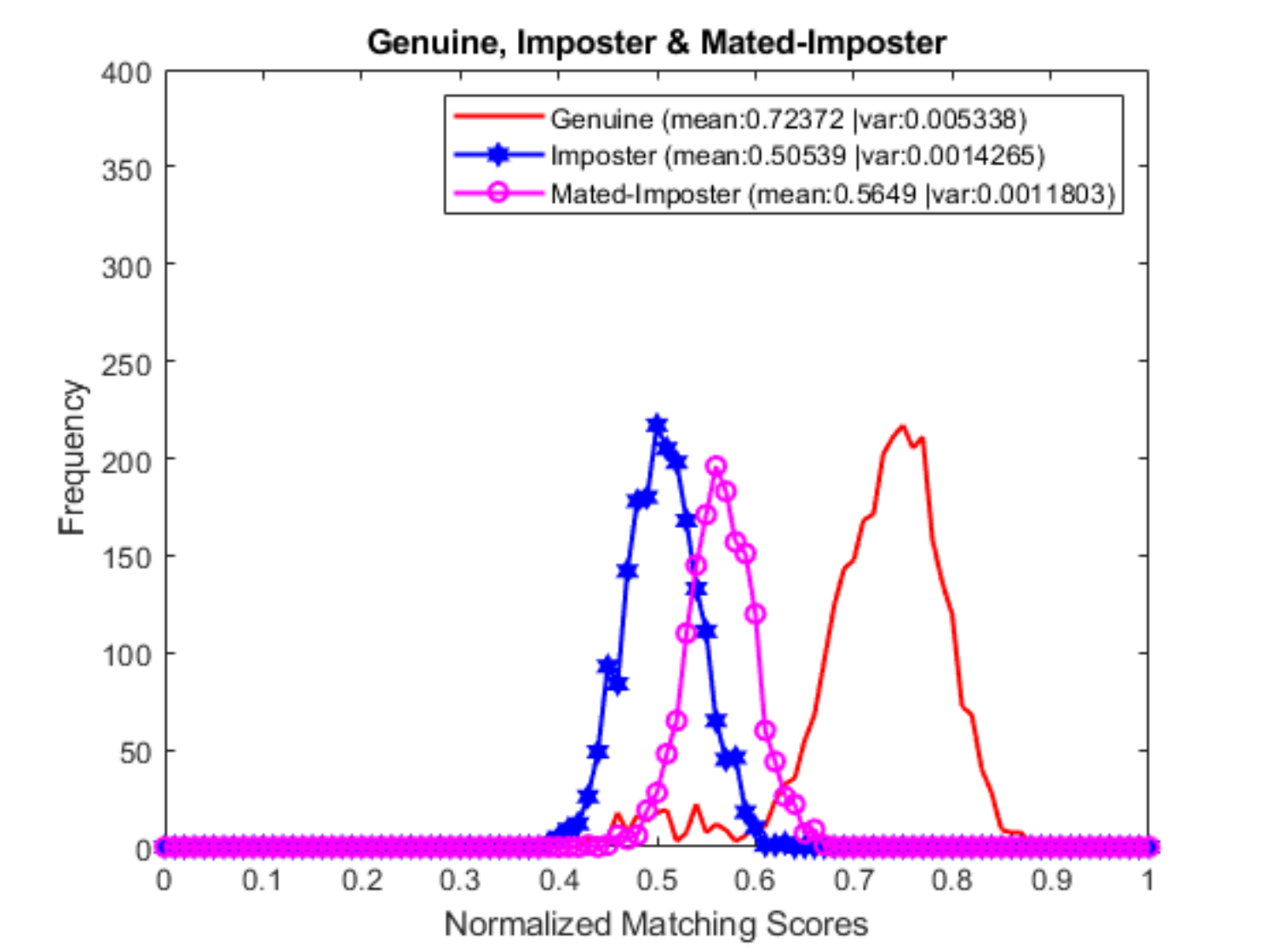}{(b) }
  \imgStub{0.32}{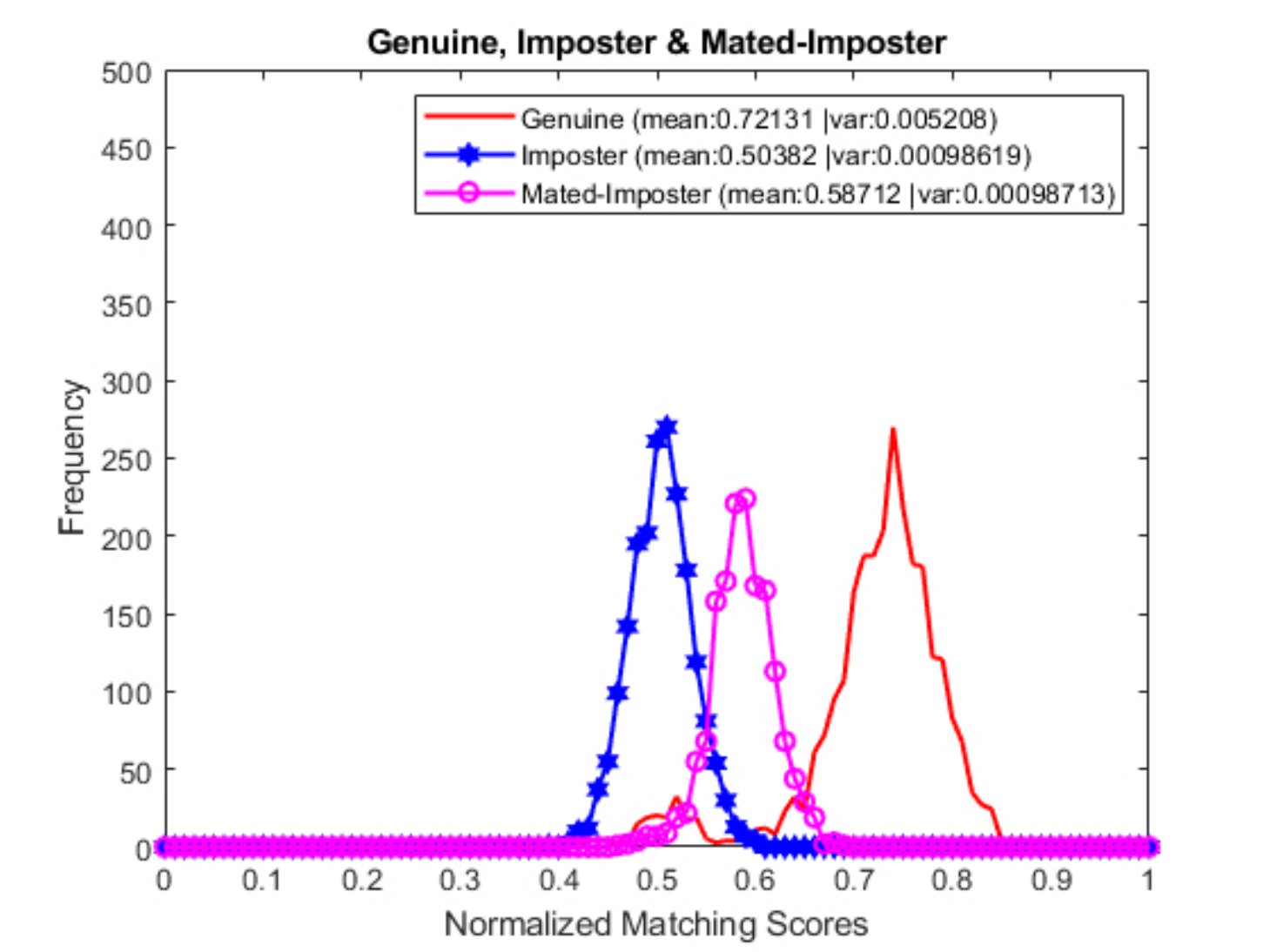}{(c) } 
  \caption{Score distribution shifting of BioHashing under attack. The bits length $l$ of (a),(b),(c) are 100,300,500 respectively.  \label{Figure::biohashing_score_distribution}}
\end{center}
\vspace{-1.0em}
\end{figure*}

In this work, the resistance to SA for BioHashing template $h(x)$ with different bit size $l$ is evaluated. Firstly, the EER, FAR of BioHashing system under helper data stolen scenario (all users use the same helper data), denoted as $sys_1$ is setup. The system threshold $\theta$ is computed and fixed with respect to the system EER under normal situation. Next, the SA is launched in $sys_2$ to find each user's pre-image (deep features) $\hat{x}$ from a transform template stored in a compromised system $sys_1$. The $h(\hat{x})$ is then used as a mated-SA-imposter to match with $h(x)$ in $sys_2$ to generate the mated-SA-imposter scores. Finally the FAR@ET under the threshold $\theta$, and the overlapping area $OL$ of imposter and mated-SA-imposter distribution are reported (see Table \ref{table::biohashingresult}, Figure \ref{Figure::biohashing_score_distribution}).

\begin{table}[t!]
\centering
\caption{Similarity-based attack evaluation on BioHashing (\%). \label{table::biohashingresult}}
\begin{tabular}{|c|c|c|c|c|c|}
\hline
 & \multicolumn{3}{c|}{Normal} & \multicolumn{2}{c|}{Attack} \\ \hline
\begin{tabular}[c]{@{}c@{}}Bits\\  length $l$ \end{tabular} & \begin{tabular}[c]{@{}c@{}} \small{Threshold} \\ $\theta$ \end{tabular} & EER & \begin{tabular}[c]{@{}c@{}}FAR\\ @ET\end{tabular} & \begin{tabular}[c]{@{}c@{}}FAR\\ @ET\end{tabular} & OL \\ \hline
16 & 0.56 & 19.95 & 24.12 & 28.28 & 94.45\\ \hline
32  & 0.59 & 12.76 & 12.48 & 14.22 & 94.18\\ \hline
64  & 0.59 & 7.80  & 7.11  & 11.35 & 86.74 \\ \hline
100 & 0.58 & 6.56  & 6.81  & 14.33 & 78.73\\ \hline
200 & 0.57 & 5.49  & 5.43  & 28.78 & 53.93 \\ \hline
300 & 0.56 & 5.34  & 5.18  & 51.62 & 38.52 \\ \hline
400 & 0.56 & 5.48  & 5.57  & 72.20 & 28.38\\ \hline
500 & 0.56 & 5.29  & 5.22  & 85.54 & 19.22 \\ \hline
\end{tabular}
\end{table}

The results reveal several interesting points:
\begin{itemize}

\item  For a BioHashing secured biometric system, the longer BioHash code can achieve better accuracy performance in terms of EER and FAR@ET. 

\item However, under SA, longer BioHash code leads to poor FAR@ET, which implies weak resistance to the similarity-based attack. This is not surprising as the longer BioHash code suggests contain more information, thus can bring decent accuracy performance. However, more information implies lower attack complexity. 

\item As shown in Figure \ref{Figure::biohashing_score_distribution}, the longer BioHash code will cause larger shifting of the mated-SA-imposter distribution, and this leads to more instances which can exceed the threshold and gain access to the system. 
\end{itemize}

It is worth to point out that short BioHash code does not imply better security, since it is vulnerable to brute-force inversion attack. Thus both long and short templates are not recommended. 

\subsection{Bloom-filter}
In this paper, the Bloom-filter is implemented as described in~\cite{rathgeb2013bloomfilteriris} without using the key, since we focus on irreversibility instead of revocability. The IrisCode is evaluated by Equal Error Rate (EER). We follow the settings in ~\cite{rathgeb2013bloomfilteriris} and the EER of $\pm $8 bits shifting and direct matching of IrisCode is calculated. In our experiment, the EER of $\pm $8 bits shifting matching is 3.46\% while the direct Hamming matching is 21.6\%. 

Bloom-filter has two parameters, i.e. word size $\omega$ and block size $l$. In this work we fix the block size as $2^6$, and the word size rangers from 8 to 10 bits. We consider two SA variants:

\begin{itemize}
\item Attack 1-template: SA attack based on one compromised template per subject.
\item Attack $n$-templates: SA attack based on multiple compromised templates per subject ($n=3$ templates in this work), i.e. $\hat{x}_i \gets \{h(x_i^j)\} $, where $i$ denotes $i$-th subject, $j=1,2,3$. The fitness is calculated as $s \approx mean(d(h(\hat{x}),h(x_i^j))) $.

\end{itemize}
Like BioHashing, the system threshold $\theta$ is computed and fixed with respect to the system EER under normal situation first. Next, two SAs are launched to reconstruct back the IrisCode. The experiment is reported in Table~\ref{table::bfresult}. Firstly we can find that the SA attack can increase the FAR@ET under Attack 1-template. Unsurprisingly, Attack $n$-templates can significantly increase the FAR@ET compared with 1-template. This is due to more information are presented.Overall, we can conclude it is feasible to perform SA attack on Bloom-filter.
 
\begin{table*}[t!]
\centering
\caption{Similarity-based attack evaluation on Bloom-filter (\%) with block size $2^6$. \label{table::bfresult}}
\begin{tabular}{|c|c|c|c|c|c|c|c|}
\hline
 & \multicolumn{3}{c|}{Normal} & \multicolumn{2}{c|}{Attack 1-template} & \multicolumn{2}{c|}{Attack $n$-templates, n=3} \\ \hline
\begin{tabular}[c]{@{}c@{}}Word Size\\ $\omega$ \end{tabular} & \begin{tabular}[c]{@{}c@{}}Threshold\\ $\theta$ \end{tabular} & EER & FAR@ET & FAR@ET & OL & FAR & OL \\ \hline
8 & 0.27 & 14.07 & 13.99 & 38.03 & 33.59 & 74.21 & 29.96 \\ \hline
9 & 0.19 & 13.88 & 14.05 & 17.32 & 38.85 & 52.27 & 40.40 \\ \hline
10 & 0.14 & 14.22 & 14.19 & 13.86 & 45.19 & 42.88 & 50.37 \\ \hline
\end{tabular}
\end{table*}

\subsection{Time cost}
\begin{figure}[t]
\begin{center}
  \imgStub{0.48}{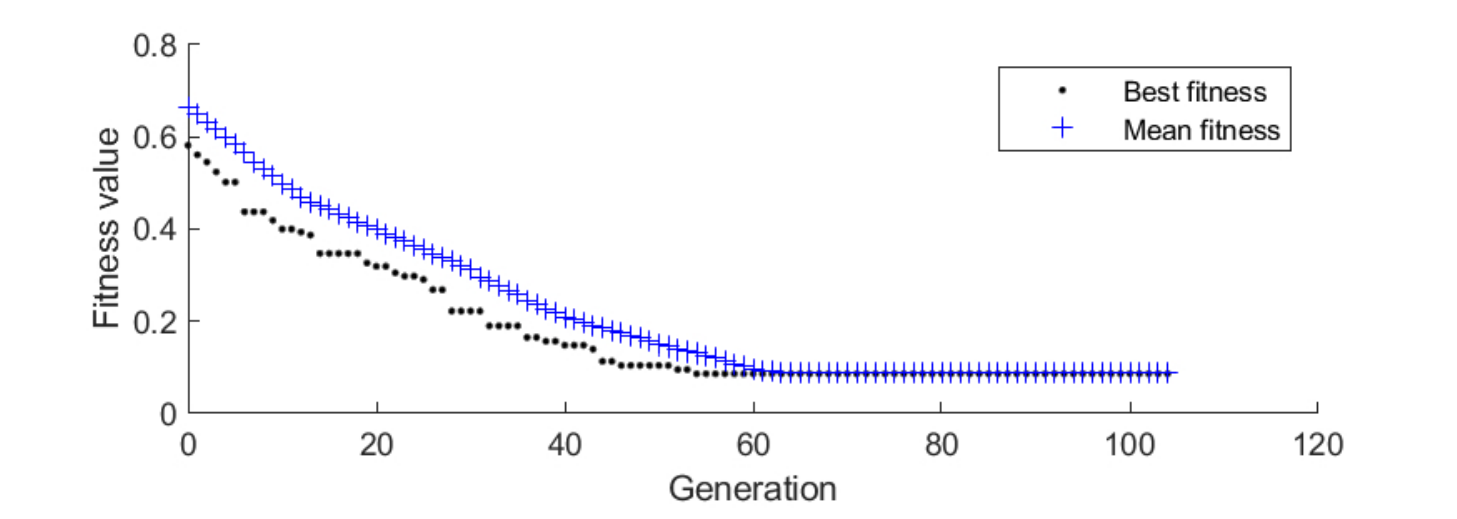}{ (a)  } \\
  \vspace{-1.0em}
  \hfil
  \imgStub{0.48}{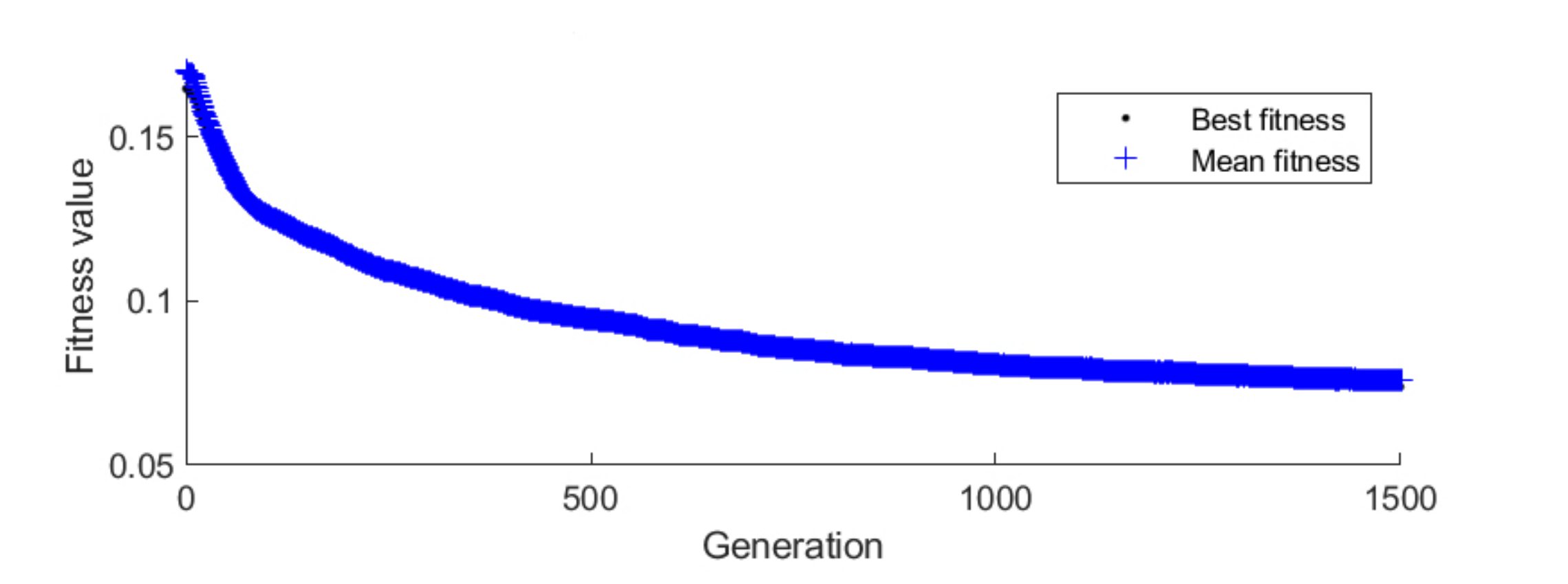}{(b)  }
  \caption{Time cost (a): BioHashing under $l=200$ bits, (b) Bloom-filter under $\omega=8$ bits.\label{Figure::time}}
\end{center}
\vspace{-2.0em}
\end{figure}

The time efficiency of the GASAF is evaluated in this section. By setting the BioHash code to 200 bits, the time spend for BioHashing are recorded. The machine we use for simulation is equipped with a MATLAB Ver. 2018b, Intel(R) Core(TM) i7-3770 CPU @ 3.40GHz and 16GB RAM. To reconstruct a face biometric feature, 35.8 seconds is spent for a template. As shown in Figure \ref{Figure::time} (a), only 50 generations is needed to find the best solution. 

As for Bloom-filter, Attack-1-template with $\omega=8$ is used to evaluate the time cost. Under this setting, it takes average 2175 seconds, 1500 generations to complete one attack (Figure \ref{Figure::time} (b)). Even though it takes around 30 minutes to complete one attack, Bloom-filter remains high risk under SA. 

In a nutshell, the result shows that the proposed attack can be time efficient and has great potential risk in real life.

\section{Conclusion}
In this work, we present a framework of Genetic Algorithm based similarity-based attack (GASAF) on cancellable biometrics. The GASAF can be launched on real-valued and binary based biometric templates.The genetic algorithm used in this work does not need training data and is time efficient. To evaluate the resistance to similarity-based attack, the overlapping area of Imposter and Mated-SA-Imposter distributions and FAR@ET are proposed. Two representative cancellable biometrics, namely BioHashing and Bloom-filter have been used as case study for this paper. 

Similarity-based attack exploits similarity preserving characteristics of cancellable biometrics to breach the biometric security while similarity preserving is a vital ingredient to keep accuracy performance intact after transformation. This suggests the trade-off between security and performance is inevitable in cancellable biometric schemes. To alleviate similarity-based attack, we recommend that the system should carefully tune the parameters to achieve a balance between security and accuracy. From this work, we notice if higher accuracy is needed, then the template must capture enough information e.g. long transformed code. However, this will lead to information leakage and weak resist to similarity-based attack. On the other hand, small code is vulnerable to brute-force inversion attack.

\section*{Acknowledgement}
This research was partly supported by Fundamental Research Grant Scheme (FRGS/1/2018/ICT02/MUSM/03/3). We also gratefully acknowledge the support of NVIDIA Corporation with the GPU grant donation of the Titan Xp GPU used for this research. Thanks for the helpful discussion with my colleague Jeffrey Ting. Special thanks for my grandfather. 
%\clearpage 
\bibliographystyle{ieeetr}
\begin{small}
\bibliography{reference}\end{small}

%-------------------------------------------------------------------------
\end{document}